\pdfoutput=1

\documentclass[10pt, a4paper]{article}

\usepackage{lrec-coling2024}

\usepackage{times}
\usepackage{latexsym}
\usepackage[T1]{fontenc}
\usepackage[latin1]{inputenc}
\usepackage{microtype}
\usepackage{inconsolata}

\usepackage{graphicx}
\DeclareGraphicsExtensions{.pdf,.ai,.jpg,.png}
\setkeys{Gin}{pagebox=artbox}

\usepackage{booktabs}
\usepackage{multirow}

\graphicspath{{./coling24-paraphrase-task-taxonomy-figures/}}

\newcommand{\mgexample}[2]{
\begin{enumerate}
\setlength{\partopsep}{0pt}
\setlength{\topsep}{0pt}
\setlength{\itemsep}{0pt}
\setlength{\parskip}{1ex}
\setlength{\parsep}{0pt}
\item[{\bf O}:] {\it #1}
\item[{\bf P}:] {\it #2}
\end{enumerate}}

\begin{document}

\title{Task-Oriented Paraphrase Analytics}

\name{Marcel Gohsen\textsuperscript{\textnormal{1}}, Matthias Hagen\textsuperscript{\textnormal{2}}, Martin Potthast\textsuperscript{\textnormal{3}}, Benno Stein\textsuperscript{\textnormal{1}}}

\address{
\textsuperscript{1}Bauhaus-Universität Weimar\\ 
\textsuperscript{2}Friedrich-Schiller-Universität Jena\\ 
\textsuperscript{3}Leipzig University and ScaDS.AI}

\abstract{
Since paraphrasing is an ill-defined task, the term ``paraphrasing'' covers text transformation tasks with different characteristics. Consequently, existing paraphrasing studies have applied quite different (explicit and implicit) criteria as to when a pair of texts is to be considered a paraphrase, all of which amount to postulating a certain level of semantic or lexical similarity. In this paper, we conduct a literature review and propose a taxonomy to organize the 25~identified paraphrasing (sub-)tasks. Using classifiers trained to identify the tasks that a given paraphrasing instance fits, we find that the distributions of task-specific instances in the known paraphrase corpora vary substantially. This means that the use of these corpora, without the respective paraphrase conditions being clearly defined (which is the normal case), must lead to incomparable and misleading results. \\[1ex]
}

\maketitleabstract

\section{Introduction}
Even though paraphrasing tasks have been studied in the field of natural language processing for decades, there is not one universally agreed ``definition'' of what exactly characterizes two texts as paraphrases of each other \cite{vila:2014}. Typical notions in the literature range from ``convey the same meaning but use different words'' \cite{bhagat:2013} over ``restatements with approximately the same meaning'' \cite{wang:2019} to ``talking about the same situation in a different way'' \cite{hirst:2003}. Such rather fuzzy and generic descriptions of the term lead to a multitude of text transformation tasks that can be subsumed under the umbrella term of paraphrasing. 

Authors from the paraphrasing literature have previously unmasked related tasks as paraphrasing tasks. \citet{zhao:2009} modified their method for paraphrase generation to solve sentence compression, sentence simplification, and sentence similarity computation. Text simplification has been referred to as ``paraphrase-oriented'' by \citet{cao:2016}. Moreover, \citet{bolshakov:2004} list several types of paraphrasing including text compression, canonization, and simplification.

We propose a novel taxonomy of 25 paraphrasing tasks, categorizing them into two groups: generating \textit{semantically equivalent} or \textit{semantically similar} paraphrases of a text. This taxonomy could pave the way for general-purpose paraphrase corpora to be used in the context of paraphrasing subtasks for which there is insufficient ground truth. Also, the taxonomy may facilitate the development of controlled paraphrase generation methods as solutions to various subtasks of paraphrasing.

The contributions of this work are threefold. First, we give a comprehensive overview of related work in the context of paraphrasing with respect to subtasks. Second, we introduce the taxonomy of paraphrasing tasks, along with their rationale and possible applications. Finally, we examine task-specific paraphrases and compute the distributions of task-specific paraphrases in general-purpose paraphrase corpora.%
\footnote{Code and data is available in the following repository: \href{https://github.com/webis-de/LREC-COLING-24/tree/main}{github.com/webis-de/LREC-COLING-24}}
As a result, we observe that the distribution of task-specific paraphrases varies substantially across multiple general-purpose paraphrase corpora. 

\section{Related Work}
\label{related-work}

We examine related work in terms of definitions, generation, and corpora of paraphrases to provide the basis for the paraphrasing task taxonomy. 

\subsection{Paraphrase Definition}
There is yet no universally accepted definition of paraphrases \cite{vila:2014}.  A common definition is that paraphrases are texts that convey the same meaning but use a different wording. 

According to \citet{bhagat:2013}, paraphrases are not always strictly semantically equivalent. In line with this hypothesis are the following definitions from the literature, which define paraphrases as sentences that have approximately the same \cite{wang:2019} or a similar meaning \cite{sun:2012}. Other works characterize paraphrases even less strictly, for example as sentences appearing in a similar context \cite{barzilay:2001}. \citet{hirst:2003} said that to paraphrase is to speak differently about the same situation. According to \citet{murata:2001}, paraphrasing comprises transforming sentences from difficult to simple or from poor to polished.

Previous works have been exclusively devoted to the question of what exactly a paraphrase is \cite{al-ghidani:2018,bhagat:2013,vila:2014}. As a result, these works usually present typologies that classify paraphrases in terms of various linguistic properties.

\subsection{Paraphrase Typology}

Paraphrases have been considered from a lexical and structural perspective, i.e., in terms of changes at the word level and syntactic level, respectively \cite{bhagat:2009,fujita:2005}. To examine these changes, paraphrase types have been distinguished according to changes at the surface level and semantic level \cite{dutrey:2011}. A finer classification of textual changes between original and paraphrased text are four classes of textual changes, namely morpholexicon-based, structure-based, semantics-based, and miscellaneous changes with several subtypes in each of these classes \cite{vila:2014}. \citet{dras:1999} classified paraphrases by their effects (e.g., loss of meaning) and introduced a typology comprising change of perspective, change of emphasis, change of relation, deletion, and clause movement.

\subsection{Paraphrase Generation}

Approaches to paraphrase generation have been primarely developed without a specific task in mind, early works with rule-based approaches \cite{barzilay:2003}, later with statistical machine translation \cite{wubben:2010,sun:2012} and recently with (deep) neural models \cite{prakash:2016,gupta:2018,li:2018,egonmwan:2019,qiu:2023,qian:2019} as highlighted in comprehensive surveys \cite{zhou:2021}. These ``general-purpose'' paraphrasing models are motivated by being beneficial for various downstream tasks for which their system is rarely tested. 

Some paraphrasing models have control mechanisms to affect syntax \cite{goyal:2020,kumar:2020} or lexical novelty \cite{chowdhury:2022}. This is a step towards task-oriented paraphrasing. However, there is a gap between controllable features and task requirements.

A few paraphrase models have been evaluated in the domain of a subtask of paraphrasing. \citet{murata:2001} have evaluated their paraphrasing model for question answering, sentence compression, and sentence polishing. Similarly, \citet{zhao:2009} have evaluated their system for sentence compression, simplification, and similarity computation. \citet{bolshakov:2004} have experimented with paraphrase generation in the area of text compression, canonicalization, and simplification. \citet{cao:2016} have analyzed their generated paraphrases in the context of text simplification and summarization.

Paraphrase generation models have been created in dedication to a specific subtask of paraphrasing including adversarial example generation \cite{iyyer:2018}, the algebraic word problem \cite{gupta:2023}, automatic evaluation \cite{kauchak:2006}, data augmentation \cite{lu:2023}, information disguise \cite{agarwal:2023}, plagiarsm detection \cite{wahle:2022} and question answering \cite{mckeown:1983}.

\subsection{Paraphrase Corpora}

Similar to paraphrase generation, most paraphrase corpora have been created as  ``general-purpose'' paraphrase collections. Of these, one commonly used dataset is the Microsoft Research Paraphrase Corpus (MSRPC), which contains 5,801 manually annotated sentence pairs from parallel corpora. A larger instance is the ParaNMT-50M dataset \cite{wieting:2018} with 50 million sentence pairs obtained by machine translation. Even larger is the Paraphrase Database (PPDB) \cite{ganitkevitch:2013, ganitkevitch:2014, pavlick:2015}, which contains more than 100 million paraphrase pairs in 23 different languages from parallel corpora. TaPaCo \cite{scherrer:2020}, also multilingual, is a paraphrase dataset with nearly 2 million sentence pairs in 73 languages, also collected from parallel corpora.

A popular method for paraphrase dataset creation is pivoting (i.e., using a pivot medium to identify semantically equivalent texts) which has been used for the creation of the Twitter URL dataset \cite{lan:2017} and the Wikipedia-IPC dataset \cite{gohsen:2023}. The former has been created by linking tweeds that contain the same url which cumulated to around 50,000 paraphrase pairs. The latter linked divergent image captions of the same image on Wikipedia which resulted in close to a million paraphrase pairs.

Unlike the aforementioned corpora, some paraphrase datasets have been created for a specific subtask of paraphrasing. One of these subtasks is plagiarism detection, for which the P4P corpus \cite{barron-cedeno:2013} has been created, based on the PAN-CPC-10 dataset \cite{potthast:2010}. Another example is the MPC corpus \cite{wahle:2022a}, which has been compiled from arXiv publications, dissertations, and Wikipedia articles.

A common task for which paraphrase corpora have been created is the identification of question duplicates in online forums. The most famous example is Quora Question Pairs \cite{quora:2017}, which contains about 400,000 potential question duplicates. In addition, \citet{fader:2013} published a paraphrase dataset for this task with 18 million paraphrase pairs collected by WikiAnswers\footnote{\url{http://wiki.answers.com/}}.

\section{Paraphrasing Task Taxonomy}
\label{task-taxonomy}

\begin{figure*}[t]
\includegraphics{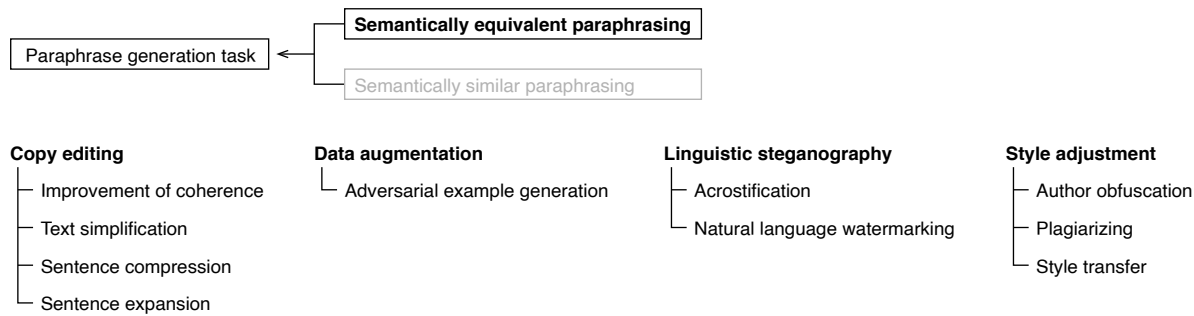}
\caption{Taxonomy of paraphrase generation tasks which require generated paraphrases to be semantically equivalent to the original text.}
\label{paraphrase-task-taxonomy1}
\end{figure*}

In this section, we disentangle paraphrasing tasks and construct a taxonomy that categorizes them as either generating semantically equivalent or similar paraphrases. The taxonomy of paraphrasing tasks has been created based on a comprehensive systematic literature review. A task is considered paraphrasing if (1) it is explicitly linked to paraphrasing by the authors or (2) its definition represents a specialization of the paraphrasing definition.

\subsection{\hspace*{-1ex}Semantically Equivalent Paraphrasing}

Semantically equivalent paraphrasing means to rewrite a text using different words so that it has exactly the same meaning as the original text. Figure~\ref{paraphrase-task-taxonomy1} provides an overview of all semantically equivalent paraphrasing tasks.

\paragraph{Copy Editing}
Copy editing is the task of rewriting text to ``remove any obstacles between the reader and what the author wants to convey'' \cite{butcher:1975}. These ``obstacles'' include spelling and grammar mistakes, repetition, ambiguity, factual errors and misleading information. Edits to overcome these obstacles that preserve the meaning of the original text are called paraphrases~\cite{faigley:1981,daxenberger:2012,yang:2021}. According to a taxonomy of edits \cite{faigley:1981}, grammar and spelling corrections are no paraphrasing tasks, even though they are meaning-preserving.

\paragraph{Improvement of Coherence}

Text coherence can be defined as ``continuity of senses'' \cite{debeaugrande:1981}, meaning that a reader can easily move across sentences and reads a paragraph as an integrated whole. A text with improved coherence should convey the same information than the original. For example, the following pair of texts from \citet{ainsworth:2007} shows a less coherent original and a more coherent paraphrase.

\mgexample{In the lungs, carbon dioxide leaves the circulating blood and oxygen enters it.}{In the lungs, carbon dioxide that has been collected from cells as blood has passed around the body, leaves the circulating blood and oxygen enters it.}

\paragraph{Text Simplification}

In text simplification, the goal is to rewrite the text using simpler grammar and words while preserving meaning. Preserving meaning makes text simplification a paraphrasing task with the added constraint of improving the readability of the original text. The following text pair is an example from the ASSET text simplification dataset \cite{alva-manchego:2020} for an original text and its simplified paraphrase.

\mgexample{He settled in London, devoting himself chiefly to practical teaching.}{He lived in London. He was a teacher.}

\noindent Text simplification has been approached as a paraphrasing problem in the literature \cite{cao:2016,xu:2016,zhao:2009}. Moreover, paraphrase corpora or generation approaches form the basis for many text simplification methods \cite{maddela:2021,yimam:2018}.

\paragraph{Sentence Compression and Expansion}

Sentence compression is about creating a ``shorter paraphrase of a sentence'' \cite{filippova:2015}. The meaning of the original sentence should be preserved. Creating a concise text that has approximately the same meaning as the original text is paraphrasing \cite{murata:2001}. Below is an example of an original text and its compression from the work of \citet{cohn:2008}.

\mgexample{The future of the nation is in your hands.}{The nation's future is in your hands.}

\noindent Vice versa, sentence expansion means paraphrasing a short sentence and expanding it in a creative way \cite{safovich:2020}.


\paragraph{Data Augmentation}

Data augmentation means generating synthetic labeled data through class label-preserving transformations \cite{kumar:2019}. In the context of tasks that require semantic equivalence (e.g., machine translation), generated examples retain the meaning of the original.

Augmenting training or test data with paraphrasing has been shown to be useful for dialog systems \cite{coca:2023,gao:2020}, machine translation \cite{callison-burch:2006,kauchak:2006,madnani:2007,madnani:2008,owczarzak:2006}, question answering \cite{dong:2017, fader:2013, fader:2014}, reading comprehension \cite{yu:2018}, summarization \cite{rush:2015}, and text classification \cite{zhang:2016, wang:2015}.

\paragraph{Adversarial Example Generation}

Adversarial examples are label-preserving modifications of texts, for which the model prediction changes \cite{szegedy:2014}. Adversarial example generation for text classification tasks has been done with paraphrase generation by \citet{iyyer:2018}, who created adversarial examples for sentiment classification and textual entailment detection. Below is an adversarial example for sentiment classification, where the original text is correctly labeled with a negative sentiment, but the paraphrased text is incorrectly classified with a positive sentiment.

\mgexample{There is no pleasure in watching a child suffer. $\rightarrow$ \normalfont{negative sentiment}}{In watching the child suffer, there is no pleasure. $\rightarrow$ \normalfont{positive sentiment}}

\paragraph{Linguistic Steganography}

The task of hiding a message in a cover signal in a way that an eavesdropper does not realize that a communication takes place is called steganography \cite{ziegler:2019}. Linguistic steganography uses textual cover signals and has been implemented with paraphrase generation \cite{chang:2010}. \citet{wilson:2014} specify that the required text transformation is equivalent to paraphrasing.

\paragraph{Acrostification}

An acrostic is a message in a text that can be decoded by concatenating the initial letters of each line. Acrostification is the task of rewriting a text such that it contains an acrostic, which was modeled by \citet{stein:2014} as a paraphrasing task. The following original text is rewritten to contain the acrostic ``HOPE''.

\mgexample{To achieve your dreams, stay optimistic and persistent despite doubts. Embrace high expectations and let your light shine.}{
\textbf{H}old onto your dream while mindful of time\\
\textbf{O}ptimism required, let your light shine\\
\textbf{P}ersistence prevails, while it may cast doubt\\
\textbf{E}xpectation desired is what it's about.}

\paragraph{Natural Language Watermarking}

A watermark in natural language is a hidden pattern in a text that is imperceptible to humans and makes it possible to identify the original author. The proof of the presence of a watermark is evidence that the text was written by the author who inserted the watermark. \citet{topkara:2005} say that paraphrasing is directly related to natural language watermarking, since it involves the modification of parameters such as length, readability or style but is intended to preserve meaning.

\paragraph{Style Adjustment}

Each text conveys characteristics of an author and is adjusted to a particular time, place and scenario \cite{jin:2022}. These characteristics are called style and are distinct from the semantic content. Style includes emotion, humor, politeness, formality, and code-switching \cite{xu:2021}. The style adjustment task aims to modify a text and control these attributes while preserving the meaning, which makes it a paraphrasing task.

\paragraph{Author Obfuscation}

The task of author obfuscation is to paraphrase a text such that the original author of that text can no longer be verified. In order to obfuscate the author, the stylistic features of the original text needs to be changed. 

In the following, we give an example of the author obfuscation approach of \citet{bevendorff:2020}. This example is an excerpt from ``Victory'' by Lester del Rey, in which the author of the original text has been obfuscated. 

\mgexample{Three billion people watching the home fleet take off, knowing the skies were open for all the hell that a savage enemy could send!}{Three billion people watching the home fleet take off, deciding the skies were resort for all the mischief that a savage enemy could send!}

\paragraph{Plagiarizing}

Plagiarism is the reuse of another person's ideas, results, or words without crediting the original author \cite{anderson:2011}. Paraphrasing is the underlying mechanism for plagiarizing text \cite{barron-cedeno:2013}. 

Below is an example from the P4P corpus \cite{potthast:2010} for an original text and its plagiarized counterpart.

\mgexample{``What a darling'' she said; ``I must give her something very nice''}{``Oh isn't she sweet!'' she said, thinking that she should present with some kind of special gift. }

\paragraph{Style Transfer}

Text style transfer is defined as changing the style of a given text without altering its semantics, which implies that style transfer is a paraphrasing task \cite{krishna:2020}.

The following text pairs represent an original text in the style of a tweet  transferred to the style of Shakespear produced by the STRAP style transfer system \cite{krishna:2020}.

\mgexample{Yall kissing before marriage?}{And you kiss'd before your nuptial?}

\subsection{Semantically Similar Paraphrasing}

\begin{figure*}[t]
\includegraphics{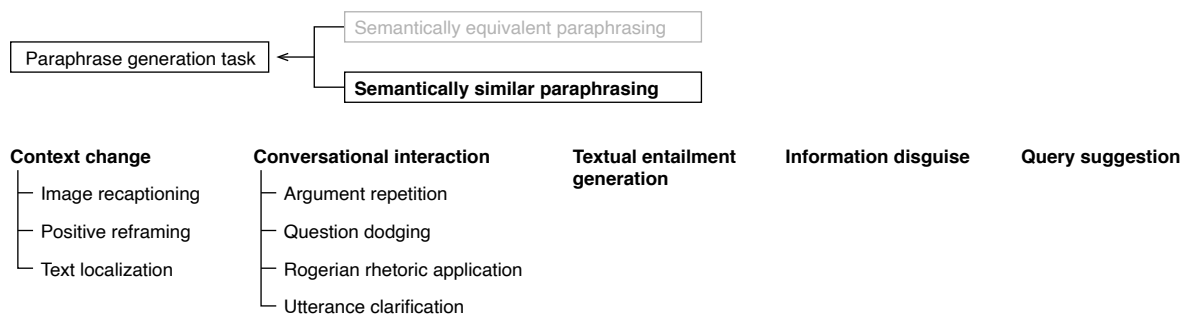}
\caption{Taxonomy of paraphrase generation tasks which allow generated paraphrases to be semantically similar and not necessarily identical to the original text.}
\label{paraphrase-task-taxonomy2}
\end{figure*}

Paraphrases are not strictly semantically equivalent \cite{bhagat:2013}, meaning that subtle semantic changes to a text preserve the paraphrase relation. We refer to these tasks as \textit{semantically similar paraphrasing}. Figure~\ref{paraphrase-task-taxonomy2} provides an overview of tasks from this category.

\paragraph{Context Change}

The background of a composition and the parts that precede and follow a given text are called context \cite{ben-amos:1993}. The paraphrasing task of context change is to rewrite a text to fit a newly given context while retaining most of its meaning.

\paragraph{Image Recaptioning}

Image recaptioning is the process of assigning a caption to an already captioned image to fit it into a new desired context. \citet{gohsen:2023} have analyzed several captions for an image and found that they are often paraphrases of each other. The popularity of MSCOCO \cite{lin:2015} (i.e., a dataset with multiple captions per image) as a training or test set for paraphrases suggests that reformulating a caption is a paraphrase generation task.

The following example is a caption pair from the Wikipedia-IPC dataset \cite{gohsen:2023} that represents paraphrases.

\mgexample{Twelfth century illustration of a man digging.}{An English serf at work digging, circa 1170.}

\noindent Because of the changing context in which the image was used (i.e., images from the Wikipedia articles on digging and on English agriculture in the Middle Ages), we can detect slight semantic changes. For example, from the paraphrase we learn that the digging man in the image is an English serf, which is not implied in the original.

\paragraph{Positive Reframing}

Positive reframing is a subtask of sentiment transfer. Sentiment transfer aims to rewrite a text in such a way that the original negative sentiment is transformed into a positive sentiment or vice versa. In contrast to sentiment transfer, a positively reframed text implies the original intention by taking a complementary positive point of view \cite{ziems:2022}. An example from the above work is used to illustrate this task.

\mgexample{This was a bland dish.}{I've made dishes that are much tastier than this one.}

\noindent The paraphrased text still conveys the original intent, but shifts the emphasis to a positive, self-affirming perspective. Because it closely follows the original, this can be considered a paraphrase. 

\paragraph{Text Localization}

Localization is the adaptation of a text to a different audience, which include groups from different regions, cultures, or ages. 

\mgexample{The price for a pound of rice is around one dollar.}{The price for half a kilo of rice is around one euro and 50 cents.}

\noindent The above example can be interpreted as paraphrasing the original for a European audience. Since a pound is not exactly half a kilo and the prices are adapted to the respective region, these texts are not exactly semantically equivalent, but similar enough to be considered paraphrases.

\paragraph{Conversational Interaction}

Paraphrasing is a natural part of human dialogue. Aspects of human communication that are paraphrasing tasks are repetition of arguments, dodging questions, use of Rogerian rhetoric, and utterances clarification.

\paragraph{Argument Repetition}

In \textit{argumentum ad nauseam}, it is assumed that an argument becomes more convincing if it is repeated over and over \cite{gilabert:2013}. Although this belief is a logical fallacy, it is still applied in human discourse. Restating the same argument (or claim) is a paraphrasing task. However, the same argument can be applied in different contexts and therefore lead to slight changes in semantics. The following example illustrates a repeated claim presented in a single discourse about movies, which are paraphrases.

\mgexample{The movie ``Die Hard'' deserves an Oscar.}{Other films have potential, but they do not deserve an Oscar like ``Die Hard'' does.}

\paragraph{Question Dodging}

Question dodging is to use rhetorical devices to avoid answering a question while giving the unaware questioner a sense of satisfaction. \citet{bull:2003} has identified six techniques to accomplish this and one of which is to acknowledge a question without answering it.  Acknowledgement can be achieved by rephrasing the question. In the following example, the original question is dodged by paraphrasing the question.

\mgexample{How will you address the problems caused by climate change?}{It is important to take action to address problems caused by climate change.}

\noindent Due to the transition from question to statement, strict semantic equivalence is never achieved. However, the main message is preserverd such that we consider this task a paraphrasing task.

\paragraph{Application of Rogerian Rhetoric}

In Rogerian rhetoric, as opposed to taking one's own position to argue against another's position, another's position is paraphrased to emphasize the strong points in the argument \cite{young:1970}. 

For the following example, imagine that the original text is an argument of the opposition and the author of the paraphrase shows that he or she respects the position by emphasizing the strong points of the argument.

\mgexample{Gun control laws is not the best solution. Agreeing that pursuing responsible gun ownership is a step in the right direction so that we reduce the number of accidents.}{Many people would agree that a key element to gun ownership is dependent upon being a responsible owner.}

\paragraph{Utterance Clarification}

Clarification requests in human dialogue have several causes. One of them is clarification of acoustic understanding \cite{schlangen:2004}. In this case, a speaker paraphrases what he or she has previously said to facilitate the interlocutor's understanding. In the following example, we have the original utterance, an interjection from someone asking for clarification, and the utterance clarification, which is a paraphrase of the original utterance. 

\begin{enumerate}
\setlength{\partopsep}{0pt}
\setlength{\topsep}{0pt}
\setlength{\itemsep}{0pt}
\setlength{\parskip}{1ex}
\setlength{\parsep}{0pt}
\item[{\bf O}:] {\it Mom said that she will pick us up at 5pm.}
{\color{gray}\item[{\bf R}:] {\it What did you say?} }
\item[{\bf P}:] {\it I said that Mom will get us at 5pm.}
\end{enumerate}

\paragraph{Textual Entailment Generation}

Textual entailment is a relationship between two texts in which one text implies the other \cite{korman:2018}. Paraphrasing can be considered as bidirectional entailment and its methods are often similar \cite{androutsopoulos:2010}. Therefore, we argue that the generation of a text which is entailed in an original is a paraphrasing task.

\paragraph{Information Disguise}

Information disguise is the task of rewriting texts such that the origin of the original cannot be determined, even with a search engine. \citet{agarwal:2023} present this task in the context of social media posts about sensitive topics (e.g., mental health, drug use) that should be made public, and present their paraphrasing method for solving this problem. 

\paragraph{Query Suggestion and Expansion}

The goal of query suggestion is to generate similar search queries for an input query to a search engine. The proposed search queries should retain the original search intent \cite{sordoni:2015}, which can be roughly equated with similar meaning. For example, when the following original query is entered into Google, a paraphrased query is suggested.

\mgexample{why do we yawn}{why do we yawn so much}
\section{Rationale of the Task Taxonomy}
\label{rationale}

Recent paraphrasing models have shown that they can effectively control syntax \cite{goyal:2020,kumar:2020} or lexical diversity \cite{chowdhury:2022} of generated paraphrases. Incorporating the paraphrasing task taxonomy (and the constraints that come with the paraphrasing subtasks) to build a controllable model to effectively generate paraphrases for each task-domain could be an important application. 

Using all the acquired knowledge about paraphrase generation may help to solve less studied or more difficult subtasks. For example, positive reframing has been recently introduced \cite{ziems:2022} and lacks sufficient training and test data.  

The taxonomy of paraphrase tasks may encourage researchers to use general-purpose paraphrase datasets as a starting point for training or evaluating paraphrase subtasks for which sufficiently large datasets are not available. For example, text style transfer is a broad problem considering all possible styles into which a text can be transferred. However, there are few publicly available resources. Deriving styles from general-purpose paraphrase datasets and using them as a style transfer dataset would solve this problem, and consequently could improve the effectiveness of automatic approaches.

\section{Paraphrase Task Classification}
\label{paraphrase-analytics}

Paraphrase corpora are usually treated as a homogenous body of paraphrases in the literature. However, indicated by the number of tasks in the taxonomy, paraphrases are rather heterogenous and the concept of a paraphrase too broad. To investigate how dissimilar paraphrase pairs from different corpora actually are, we conduct a blind test with a human annotator to assign the associated paraphrase subtasks to task-specific paraphrase examples. Then, we automatically classify the associated tasks to examine the distributions of task-specific paraphrases in general-purpose paraphrase datasets.

\subsection{Task-Specific Paraphrase Datasets}

From the paraphrasing task taxonomy, we select five subtasks for which sufficiently large training and test datasets are available: text simplification, sentence compression, style transfer, image recaptioning, and textual entailment  (three semantically equivalent and two semantically similar paraphrasing tasks). For each task, we employ two datasets to ensure some topical diversity to reduce detectability based on content.

In terms of text simplification, we use the TurkCorpus \cite{xu:2016} and the WikiLarge dataset \cite{zhang:2017}. The TurkCorpus contains 2,350 texts with eight simplifications each collected through crowdsourcing.  The WikiLarge dataset is an aggregation of Wikipedia-based text simplification corpora with 296,402 sentence pairs.

For sentence compression, we use Google's compression datataset \cite{filippova:2013} and Microsoft's abstractive compression dataset \cite{toutanova:2016}. They contain 250,000 sentence pairs from news headlines and about 26,000 pairs from the OANC\footnote{\url{https://anc.org/data/oanc/}}, respectively.

Regarding style transfer, we employ ParaDetox \cite{logacheva:2022} and a Bible style tansfer dataset \cite{carlson:2018}. The former is a dataset of more than 10,000 pairs of toxic and non-toxic texts from social media posts. The second dataset contains 1.7 million sentence pairs with 34 different styles from individual Bible versions.

For the image recaptioning task, we use the popular MSCOCO \cite{lin:2015} and the VizWiz dataset \cite{gurari:2020}. Both datasets contain multiple crowdsourced captions per image (about 1.5 million and 200,000 captions, respectively).

Since some popular textual entailment datasets originate from image captions (e.g., SNLI \cite{bowman:2015}), we have to use datasets that do not interfere with the other tasks. One of which is SciTail \cite{khot:2018}, a crowdsourced entailment dataset with 27,000 examples. The second dataset is HELP \cite{yanaka:2019} containing 36,000 automatically generated inference examples based on the Parallel Meaning Bank \cite{abzianidze:2017}. From the textual entailment datasets we only draw examples explicitly labeled as entailed since other examples are not paraphrases.

\subsection{Manual Task Annotation}

\begin{figure}[t]
\centering
\includegraphics[width=\linewidth]{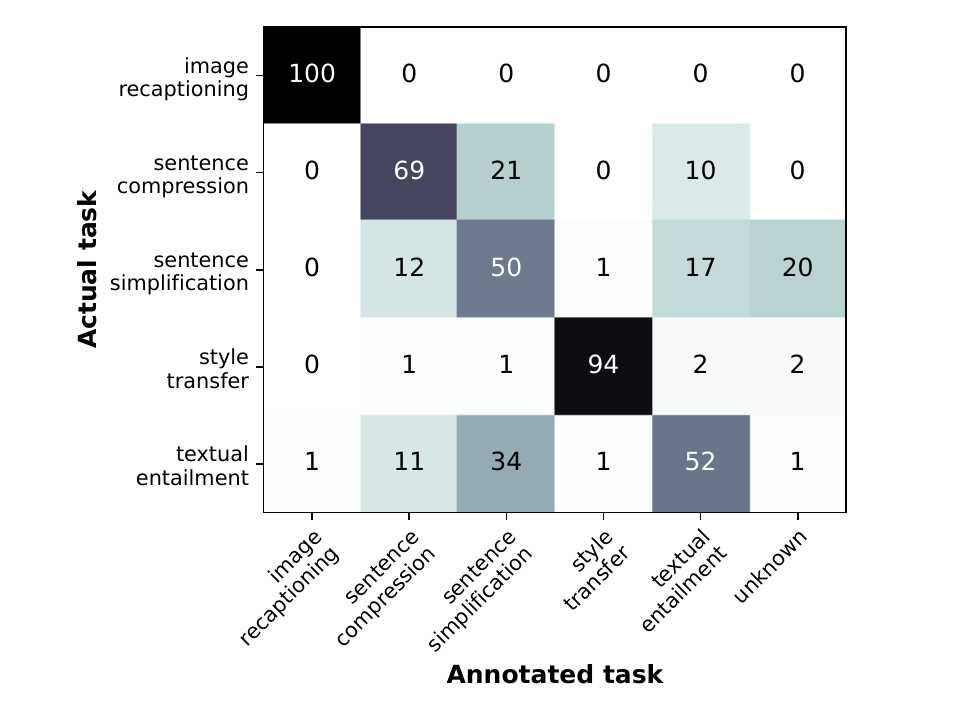}
\caption{Confusion matrix of manually annotated tasks and the actual tasks of paraphrases from task-specific corpora.}
\label{fig:paraphrase-task-confusion}
\end{figure}

To investigate whether a human can distinguish between task-specific paraphrase pairs, we conduct a manual annotation study. For that, we employ an expert annotator with over three years of experience in the field of NLP. To prepare for the annotation process, the annotator studies common definitions of all five considered paraphrasing subtasks. Given a pair of paraphrases, the annotator decides to which of the five considered paraphrasing tasks the text pair fits best. If a decision cannot be made, the annotator may assign ``unknown''.

The target of the annotation study is a total of 500 paraphrase pairs (100 examples per paraphrase subtask). Several measures are taken to mitigate bias, including (1) randomizing the order of presented paraphrases, (2) randomly selecting the same number of paraphrase pairs from each dataset per task, (3) selecting paraphrases uniformly by length, and (4) establishing a common range of paraphrase lengths for all examples, ranging from 100 to 180 total characters.

Figure \ref{fig:paraphrase-task-confusion} shows the confusion matrix of the annotated and the actual task of the 500 paraphrase pairs from task-specific corpora. Due to their descriptive and visual nature, caption pairs have been correctly identified in 100\% of the cases. It is similarly easy to identify whether text pairs have different text styles (94\% accuracy).

Distinguishing between sentence compression, simplification, and entailment examples proves to be a harder task. Examples from all three of these tasks share common signals. For example, the length delta between the texts of a paraphrase pair is a signaling indicator of compressed, entailed, and sometimes simplified sentences. Nevertheless, the annotator assigned at least 50\% of examples correctly to these three tasks. 

The majority of paraphrases for which the annotator has chosen ``unknown'' originate from corpus artifacts. For example, some text pairs do not share the same meaning at all due to misaligned sentences in the WikiLarge dataset. In these cases it is impossible to assign the corresponding task. 

The overall good accuracy with which a human can distinguish between paraphrases from different subtasks shows that the diversity of paraphrases is rather obvious and presumably observable in general-purpose paraphrase datasets, too. If this hypothesis is true, paraphrase evaluation and identification should account for this diversity. To test this, we develop a method to automatically assign one of these five paraphrase subtasks to text pairs. 

\subsection{Automatic Task Classification}

We hypothesize that the distributions of task-specific paraphrases differ substantially across different general-purpose paraphrase datasets. To test this hypothesis, we develop an automatic classifier that assigns paraphrasing tasks to a pair of texts. To enable the classifier to generalize beyond datasets within the trainig data, we focus on discriminative features that are topic-independent and operate mostly at the lexical or syntactic level.

\paragraph{Task Classifier}
To build a paraphrasing task classifier, we rely on feature engineering. As found in the annotation study, the sentence compression, simplification, and textual entailment examples stand out due to their length delta. Therefore, we use the compression ratio (i.e., the ratio between the length of the shorter and longer text) as a feature. To quantify surface-level similarity, we use ROUGE1 \cite{lin:2004} and BLEU \cite{papineni:2002}. For semantic similarity, which is crucial to distinguish task instances of semantically similar and equivalent tasks, we use the cosine similarity of Sentence-BERT embeddings \cite{reimers:2019}. Finally, we append the vectorized relative frequencies of POS tag n-grams (up to 4-grams) to the feature vector to represent the syntactical structure of the paraphrases.

We randomly sample 50,000 task-specific paraphrases (10,000 per task) and divide them into a training and a test set, maintaining a 80:20 ratio, and ensuring an even distribution of tasks. In previous experiments, we have found that a Random Forest classifier performed best on our data but also has the problem of overfitting on our data which we reduce by limitting the depth of each decision tree to 15. The effectiveness of our multi-class classifier is evaluated in a 5-fold cross-validation. The classifier achieves a micro-averaged F1 of 0.82 in the cross-validation and an F1 of 0.81 with respect to the original train-test split.

\begin{figure}[t]
\centering
\includegraphics[width=\linewidth]{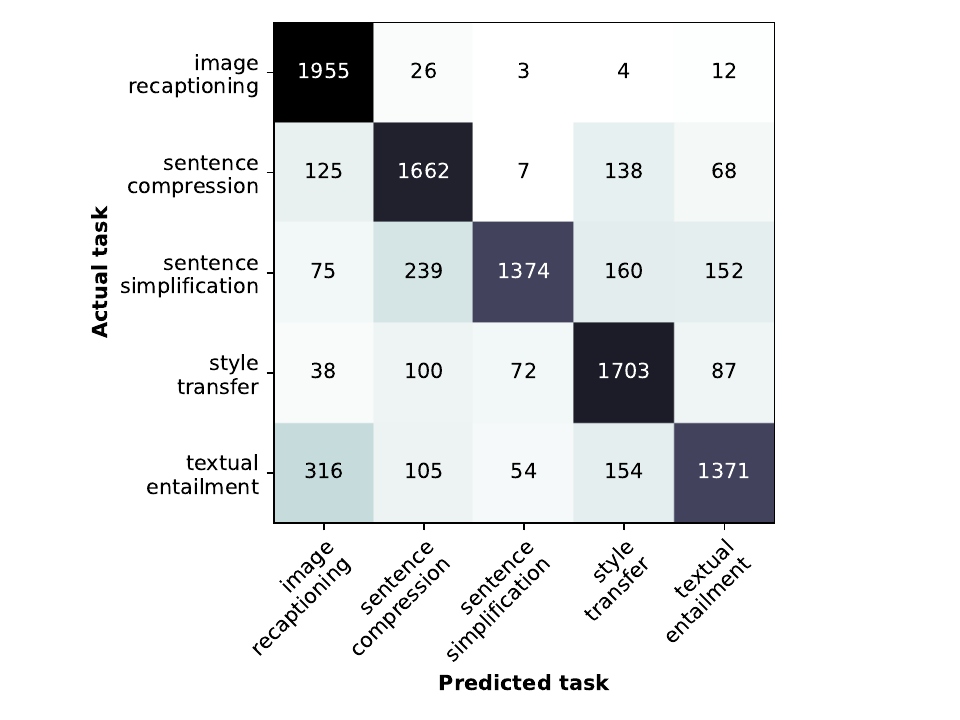}
\caption{Confusion matrix of the automatically predicted paraphrasing tasks and the actual tasks of paraphrases from task-specific corpora in the sampled test-set.}
\label{paraphrase-task-prediction-confusion}
\end{figure}

\begin{table*}[t]
\fontsize{8pt}{10pt}\selectfont
\centering
\setlength{\tabcolsep}{5pt}
\begin{tabular}{@{}l rr rr rr rr rr | r@{}}
\toprule
\multirow{2}{*}{\bf {\shortstack[l]{Paraphrase\\Dataset}}} & \multicolumn{2}{c}{\multirow{2}{*}{\bf{\shortstack[c]{Image\\Recaptioning}}}} & \multicolumn{2}{c}{\multirow{2}{*}{\textbf{\shortstack[c]{Sentence\\Compression}}}} & \multicolumn{2}{c}{\multirow{2}{*}{\textbf{\shortstack[c]{Sentence\\Simplification}}}} & \multicolumn{2}{c}{\multirow{2}{*}{\textbf{\shortstack[c]{Style\\Transfer}}}} & \multicolumn{2}{c}{\multirow{2}{*}{\textbf{\shortstack[c]{Textual\\Entailment}}}} & \multirow{2}{*}{\textbf{\shortstack[c]{Total}}}\\
\addlinespace[1em]
\midrule
MSRPC & 6.7\% & \color{gray} 390 & 32.0\% & \color{gray} 1,858 & 38.6\% & \color{gray} 2,241 & 11.4\% & \color{gray} 653 & 11.4\% & \color{gray} 659 & 5,801\\
PAWS & 5.2\% & \color{gray} 3,367 & 24.7\% & \color{gray} 16,194 & \bf 62.7\% & \color{gray} 41,004 & 3.7\% & \color{gray} 2,442 & 3.6\% & \color{gray} 2,394 & 65,401\\
TaPaCo & 1.8\% & \color{gray} 4,140 & 8.4\% & \color{gray} 18,949 & 1.0\% & \color{gray} 2,141 & \bf 76.8\% & \color{gray} 172,718 & \bf 12.0\% & \color{gray} 26,877 & 224,825\\
Wikipedia-IPC$_{\mathrm{silver}}$ & \bf 16.3\% & \color{gray} 37,489 & \bf 62.0\% & \color{gray} 142,492 & 19.8\% & \color{gray} 45,535 & 0.2\% & \color{gray} 427 & 1.7\% & \color{gray} 3,934 & 229,877\\
\midrule
\bf Total & 8.6\% & \color{gray} 45,386 & 34.1\% & \color{gray} 179,493 & 17.3\% & \color{gray} 90,921 & 33.5\% & \color{gray} 176,240 & 6.4\% & \color{gray} 33,864 & 525,904\\
\bottomrule
\end{tabular}

\caption{Frequencies of predicted task-specific paraphrases in general-purpose paraphrase corpora.}
\label{table-task-specific-paraphrase-distribution}
\end{table*}

Figure~\ref{paraphrase-task-prediction-confusion} presents a confusion matrix of the automatically predicted paraphrasing tasks by our classifier and the actual underlying task from our sampled test data. This matrix reveals similar characteristics between the human annotator and the trained classifier. Both the human annotator and the classifier reliably identify image captions and confuse sentence compression, simplification and textual entailment examples. In contrast to the human annotator, the classifier have had problems to spot style transfer examples. According to Figure~\ref{fig:paraphrase-task-confusion} and Figure~\ref{paraphrase-task-prediction-confusion}, the human annotator detects style transfer with an accuracy of 94\% while the classifier only reaches an accuracy of about 85\%.

\paragraph{Paraphrase Corpora Analytics}

We use the created classifier to assign tasks to paraphrase pairs from general-purpose paraphrase corpora including the MSRPC dataset \cite{dolan:2005}, TaPaCo \cite{scherrer:2020}, PAWS \cite{zhang:2019} and the Wikipedia-IPC dataset \cite{gohsen:2023} to investigate their task bias.

In Table~\ref{table-task-specific-paraphrase-distribution} we present relative frequencies of paraphrase pairs that have been predicted to fit to the specific paraphrasing tasks. We can see that all the analyzed corpora are biased towards a different subtask. PAWS is highly skewed towards sentence simplification, TaPaCo towards style transfer, and the Wikipedia-IPC dataset towards sentence compression. Wikipedia-IPC is a paraphrase dataset compiled from image captions for which it is surprising that the identified portion of image captions is rather low. However, the average compression ratio in the Wikipedia-IPC is 0.7 while for the TaPaCo dataset it is 0.82 which could be an explanation for these findings. The MSRPC dataset is the most heterogenous dataset out of all with a slight bias towards sentence compression and sentence simplification. The least represented task in all datasets is textual entailment. 

These results imply that paraphrases from different corpora are highly diverse and should not be considered as a homogenous pool of paraphrases for evaluating paraphrase generation. Evaluating different paraphrase generation systems on different corpora may lead to incomparable results.

\section{Conclusion}

Based on an extensive literature review, we proposed a novel taxonomy of paraphrasing tasks to support future research on task-oriented paraphrase generation. Our task classification results show that substantial biases towards different paraphrasing tasks exist among general-purpose paraphrase datasets. Therefore, treating different general-purpose paraphrase datasets as a homogeneous set for evaluation leads to incomparable results in paraphrase system evaluations.

In the future, we plan to dive deeper and investigate paraphrase systems on task-specific biases. To investigate the generalizability of the task classifier, we will evaluate its performance on unseen task-specific paraphrase corpora and test its correlation with paraphrase recognition systems for different subtasks. Since we have seen that paraphrases can belong to multiple tasks, we will extend the approach to a multi-label approach.

\section{Limitations}

The generalizability of the task classifier to unseen corpora is only sparsely evaluated. The observation that a low number of image captions have been identified in the Wikipedia-IPC dataset (i.e., a paraphrase dataset composed of captions) raises concerns about generalizability. However, even if the generalizability is poor, the finding that paraphrases from different datasets vary considerably still holds. A manual annotation process of paraphrases from general-purpose datasets could be helpful to confirm our findings and to ensure the reliability of our classifier.

The classifier does not assign multiple tasks to a pair of paraphrases. The annotation study has shown that paraphrasing tasks are not always distinct. Retraining the task classifier in a multi-label fashion might shed some light on the commonalities between the different subtasks. However, obtaining multi-labeled training examples for paraphrasing tasks is substantially harder.

\newpage
\section{Bibliographical References}
\bibliographystyle{lrec-coling2024-natbib}
\bibliography{coling24-lit}

\begin{thebibliography}{108}
\expandafter\ifx\csname natexlab\endcsname\relax\def\natexlab#1{#1}\fi

\bibitem[{Abzianidze et~al.(2017)Abzianidze, Bjerva, Evang, Haagsma, {van
  Noord}, Ludmann, Nguyen, and Bos}]{abzianidze:2017}
Lasha Abzianidze, Johannes Bjerva, Kilian Evang, Hessel Haagsma, Rik {van
  Noord}, Pierre Ludmann, Duc-Duy Nguyen, and Johan Bos. 2017.
\newblock The {{Parallel Meaning Bank}}: {{Towards}} a multilingual corpus of
  translations annotated with compositional meaning representations.
\newblock In \emph{Proceedings of the 15th Conference of the {{European}}
  Chapter of the Association for Computational Linguistics: {{Volume}} 2, Short
  Papers}, pages 242--247, {Valencia, Spain}. {Association for Computational
  Linguistics}.

\bibitem[{Agarwal et~al.(2023)Agarwal, Gupta, Bonagiri, Gaur, Reagle, and
  Kumaraguru}]{agarwal:2023}
Anmol Agarwal, Shrey Gupta, Vamshi Bonagiri, Manas Gaur, Joseph Reagle, and
  Ponnurangam Kumaraguru. 2023.
\newblock Towards {{Effective Paraphrasing}} for {{Information Disguise}}.
\newblock In Jaap Kamps, Lorraine Goeuriot, Fabio Crestani, Maria Maistro,
  Hideo Joho, Brian Davis, Cathal Gurrin, Udo Kruschwitz, and Annalina Caputo,
  editors, \emph{Advances in {{Information Retrieval}}}, volume 13981, pages
  331--340. {Springer Nature Switzerland}, {Cham}.

\bibitem[{Ainsworth and Burcham(2007)}]{ainsworth:2007}
Shaaron Ainsworth and Sarah Burcham. 2007.
\newblock {The Impact of Text Coherence on Learning by Self-Explanation}.
\newblock \emph{Learning and Instruction}, 17(3):286--303.

\bibitem[{{Al-Ghidani} and Fahmy(2018)}]{al-ghidani:2018}
Ahmed~H. {Al-Ghidani} and Aly~A. Fahmy. 2018.
\newblock {Conditional Text Paraphrasing: {{A}} Survey and Taxonomy}.
\newblock \emph{International Journal of Advanced Computer Science and
  Applications}, 9.

\bibitem[{{Alva-Manchego} et~al.(2020){Alva-Manchego}, Martin, Bordes, Scarton,
  Sagot, and Specia}]{alva-manchego:2020}
Fernando {Alva-Manchego}, Louis Martin, Antoine Bordes, Carolina Scarton,
  Beno{\^i}t Sagot, and Lucia Specia. 2020.
\newblock {{ASSET}}: {{A Dataset}} for {{Tuning}} and {{Evaluation}} of
  {{Sentence Simplification Models}} with {{Multiple Rewriting
  Transformations}}.
\newblock In \emph{Proceedings of the 58th {{Annual Meeting}} of the
  {{Association}} for {{Computational Linguistics}}}, pages 4668--4679,
  {Online}. {Association for Computational Linguistics}.

\bibitem[{Anderson and Steneck(2011)}]{anderson:2011}
Melissa~S. Anderson and Nicholas~H. Steneck. 2011.
\newblock {The Problem of Plagiarism}.
\newblock \emph{Urologic Oncology: Seminars and Original Investigations},
  29(1):90--94.

\bibitem[{Androutsopoulos and Malakasiotis(2010)}]{androutsopoulos:2010}
Ion Androutsopoulos and Prodromos Malakasiotis. 2010.
\newblock A {{Survey}} of {{Paraphrasing}} and {{Textual Entailment Methods}}.
\newblock \emph{Journal of Artificial Intelligence Research}, 38:135--187.

\bibitem[{{Barr{\'o}n-Cede{\~n}o} et~al.(2013){Barr{\'o}n-Cede{\~n}o}, Vila,
  Mart{\'i}, and Rosso}]{barron-cedeno:2013}
Alberto {Barr{\'o}n-Cede{\~n}o}, Marta Vila, M.~Mart{\'i}, and Paolo Rosso.
  2013.
\newblock Plagiarism {{Meets Paraphrasing}}: {{Insights}} for the {{Next
  Generation}} in {{Automatic Plagiarism Detection}}.
\newblock \emph{Computational Linguistics}, 39(4):917--947.

\bibitem[{Barzilay and Lee(2003)}]{barzilay:2003}
Regina Barzilay and Lillian Lee. 2003.
\newblock {Learning to Paraphrase: {{An}} Unsupervised Approach Using
  Multiple-Sequence Alignment}.
\newblock In \emph{Proceedings of the 2003 Human Language Technology Conference
  of the North {{American}} Chapter of the Association for Computational
  Linguistics}, pages 16--23.

\bibitem[{Barzilay and McKeown(2001)}]{barzilay:2001}
Regina Barzilay and Kathleen~R. McKeown. 2001.
\newblock {Extracting Paraphrases from a Parallel Corpus}.
\newblock In \emph{Association for Computational Linguistic, 39th Annual
  Meeting and 10th Conference of the European Chapter, Proceedings of the
  Conference, July 9-11, 2001, Toulouse, France}, pages 50--57. {Morgan
  Kaufmann Publishers}.

\bibitem[{{Ben-Amos}(1993)}]{ben-amos:1993}
Dan {Ben-Amos}. 1993.
\newblock "{{Context}}" in context.
\newblock \emph{Western Folklore}, 52(2/4):209--226.

\bibitem[{Bevendorff et~al.(2020)Bevendorff, Wenzel, Potthast, Hagen, and
  Stein}]{bevendorff:2020}
Janek Bevendorff, Tobias Wenzel, Martin Potthast, Matthias Hagen, and Benno
  Stein. 2020.
\newblock {On Divergence-Based Author Obfuscation: {{An}} Attack on the State
  of the Art in Statistical Authorship Verification}.
\newblock \emph{it - Information Technology}, 62(2):99--115.

\bibitem[{Bhagat(2009)}]{bhagat:2009}
Rahul Bhagat. 2009.
\newblock \emph{Learning Paraphrases from Text}.
\newblock Ph.D. thesis, University of Southern Carolina.

\bibitem[{Bhagat and Hovy(2013)}]{bhagat:2013}
Rahul Bhagat and Eduard Hovy. 2013.
\newblock Squibs: {{What Is}} a {{Paraphrase}}?
\newblock \emph{Computational Linguistics}, 39(3):463--472.

\bibitem[{Bolshakov and Gelbukh(2004)}]{bolshakov:2004}
Igor~A. Bolshakov and Alexander Gelbukh. 2004.
\newblock Synonymous {{Paraphrasing Using WordNet}} and {{Internet}}.
\newblock In David Hutchison, Takeo Kanade, Josef Kittler, Jon~M. Kleinberg,
  Friedemann Mattern, John~C. Mitchell, Moni Naor, Oscar Nierstrasz,
  C.~Pandu~Rangan, Bernhard Steffen, Madhu Sudan, Demetri Terzopoulos, Dough
  Tygar, Moshe~Y. Vardi, Gerhard Weikum, Farid Meziane, and Elisabeth
  M{\'e}tais, editors, \emph{Natural {{Language Processing}} and {{Information
  Systems}}}, volume 3136, pages 312--323. {Springer Berlin Heidelberg},
  {Berlin, Heidelberg}.

\bibitem[{Bowman et~al.(2015)Bowman, Angeli, Potts, and Manning}]{bowman:2015}
Samuel~R. Bowman, Gabor Angeli, Christopher Potts, and Christopher~D. Manning.
  2015.
\newblock {A Large Annotated Corpus for Learning Natural Language Inference}.
\newblock In \emph{Proceedings of the 2015 {{Conference}} on {{Empirical
  Methods}} in {{Natural Language Processing}}}, pages 632--642, {Lisbon,
  Portugal}. {Association for Computational Linguistics}.

\bibitem[{Bull(2003)}]{bull:2003}
Peter Bull. 2003.
\newblock \emph{The Microanalysis of Political Communication: {{Claptrap}} and
  Ambiguity}, volume~7.
\newblock {Routledge}.

\bibitem[{Butcher(1975)}]{butcher:1975}
Judith Butcher. 1975.
\newblock \emph{Copy-Editing: {{The}} Cambridge Handbook.}
\newblock {ERIC}.

\bibitem[{{Callison-Burch} et~al.(2006){Callison-Burch}, Koehn, and
  Osborne}]{callison-burch:2006}
Chris {Callison-Burch}, Philipp Koehn, and Miles Osborne. 2006.
\newblock Improved statistical machine translation using paraphrases.
\newblock In \emph{Proceedings of the Human Language Technology Conference of
  the {{NAACL}}, Main Conference}, pages 17--24, {New York City, USA}.
  {Association for Computational Linguistics}.

\bibitem[{Cao et~al.(2016)Cao, Luo, Li, and Li}]{cao:2016}
Ziqiang Cao, Chuwei Luo, Wenjie Li, and Sujian Li. 2016.
\newblock Joint {{Copying}} and {{Restricted Generation}} for {{Paraphrase}}.

\bibitem[{Carlson et~al.(2018)Carlson, Riddell, and Rockmore}]{carlson:2018}
Keith Carlson, Allen Riddell, and Daniel Rockmore. 2018.
\newblock {Evaluating Prose Style Transfer with the {{Bible}}}.
\newblock \emph{Royal Society Open Science}, 5(10):171920.

\bibitem[{Chang and Clark(2010)}]{chang:2010}
Ching{-}Yun Chang and Stephen Clark. 2010.
\newblock {Linguistic Steganography Using Automatically Generated Paraphrases}.
\newblock In \emph{Human Language Technologies: Conference of the North
  American Chapter of the Association of Computational Linguistics,
  Proceedings, June 2-4, 2010, Los Angeles, California, {USA}}, pages 591--599.
  The Association for Computational Linguistics.

\bibitem[{Chowdhury et~al.(2022)Chowdhury, Zhuang, and Wang}]{chowdhury:2022}
Jishnu~Ray Chowdhury, Yong Zhuang, and Shuyi Wang. 2022.
\newblock Novelty {{Controlled Paraphrase Generation}} with {{Retrieval
  Augmented Conditional Prompt Tuning}}.
\newblock \emph{Proceedings of the AAAI Conference on Artificial Intelligence},
  36(10):10535--10544.

\bibitem[{Coca et~al.(2023)Coca, Tseng, Lin, and Byrne}]{coca:2023}
Alexandru Coca, Bo-Hsiang Tseng, Weizhe Lin, and Bill Byrne. 2023.
\newblock {More Robust Schema-Guided Dialogue State Tracking via Tree-Based
  Paraphrase Ranking}.
\newblock In \emph{Findings of the Association for Computational Linguistics:
  {{EACL}} 2023, Dubrovnik, Croatia, May 2-6, 2023}, pages 1413--1424.
  {Association for Computational Linguistics}.

\bibitem[{Cohn and Lapata(2008)}]{cohn:2008}
Trevor Cohn and Mirella Lapata. 2008.
\newblock {Sentence Compression beyond Word Deletion}.
\newblock In \emph{Proceedings of the 22nd International Conference on
  Computational Linguistics (Coling 2008)}, pages 137--144, {Manchester, UK}.
  {Coling 2008 Organizing Committee}.

\bibitem[{Daxenberger and Gurevych(2012)}]{daxenberger:2012}
Johannes Daxenberger and Iryna Gurevych. 2012.
\newblock {A Corpus-Based Study of Edit Categories in Featured and Non-Featured
  {{Wikipedia}} Articles}.
\newblock In \emph{Proceedings of {{COLING}} 2012}, pages 711--726, {Mumbai,
  India}. {The COLING 2012 Organizing Committee}.

\bibitem[{De~Beaugrande and Dressler(1981)}]{debeaugrande:1981}
Robert-Alain De~Beaugrande and Wolfgang~U Dressler. 1981.
\newblock \emph{Introduction to Text Linguistics}, volume~1.
\newblock {longman London}.

\bibitem[{Dolan and Brockett(2005)}]{dolan:2005}
Bill Dolan and Chris Brockett. 2005.
\newblock {Automatically Constructing a Corpus of Sentential Paraphrases}.
\newblock In \emph{Third International Workshop on Paraphrasing ({{IWP2005}})}.
  {Asia Federation of Natural Language Processing}.

\bibitem[{Dong et~al.(2017)Dong, Mallinson, Reddy, and Lapata}]{dong:2017}
Li~Dong, Jonathan Mallinson, Siva Reddy, and Mirella Lapata. 2017.
\newblock Learning to {{Paraphrase}} for {{Question Answering}}.
\newblock In \emph{Proceedings of the 2017 {{Conference}} on {{Empirical
  Methods}} in {{Natural}} {{Language Processing}}}, pages 875--886,
  {Copenhagen, Denmark}. {Association for Computational Linguistics}.

\bibitem[{Dras(1999)}]{dras:1999}
Mark Dras. 1999.
\newblock \emph{Tree Adjoining Grammar and the Reluctant Paraphrasing of Text}.
\newblock Ph.D. thesis, Macquarie University Sydney.

\bibitem[{Dutrey et~al.(2011)Dutrey, Bouamor, Bernhard, and Max}]{dutrey:2011}
Camille Dutrey, Houda Bouamor, Delphine Bernhard, and Aur{\'e}lien Max. 2011.
\newblock {Local Modifications and Paraphrases in {{Wikipedia}}'s Revision
  History}.
\newblock \emph{Proces. del Leng. Natural}, 46:51--58.

\bibitem[{Egonmwan and Chali(2019)}]{egonmwan:2019}
Elozino Egonmwan and Yllias Chali. 2019.
\newblock {Transformer and Seq2seq Model for {{Paraphrase Generation}}}.
\newblock In \emph{Proceedings of the 3rd {{Workshop}} on {{Neural Generation}}
  and {{Translation}}}, pages 249--255, {Hong Kong}. {Association for
  Computational Linguistics}.

\bibitem[{Fader et~al.(2013)Fader, Zettlemoyer, and Etzioni}]{fader:2013}
Anthony Fader, Luke Zettlemoyer, and Oren Etzioni. 2013.
\newblock {Paraphrase-Driven Learning for Open Question Answering}.
\newblock In \emph{Proceedings of the 51st Annual Meeting of the Association
  for Computational Linguistics (Volume 1: {{Long}} Papers)}, pages 1608--1618,
  {Sofia, Bulgaria}. {Association for Computational Linguistics}.

\bibitem[{Fader et~al.(2014)Fader, Zettlemoyer, and Etzioni}]{fader:2014}
Anthony Fader, Luke Zettlemoyer, and Oren Etzioni. 2014.
\newblock {Open Question Answering over Curated and Extracted Knowledge Bases}.
\newblock In \emph{Proceedings of the 20th {{ACM SIGKDD}} International
  Conference on Knowledge Discovery and Data Mining}, {{KDD}} '14, pages
  1156--1165, {New York, NY, USA}. {Association for Computing Machinery}.

\bibitem[{Faigley and Witte(1981)}]{faigley:1981}
Lester Faigley and Stephen Witte. 1981.
\newblock {Analyzing Revision}.
\newblock \emph{College Composition and Communication}, 32(4):400--414.

\bibitem[{Filippova et~al.(2015)Filippova, Alfonseca, Colmenares, Kaiser, and
  Vinyals}]{filippova:2015}
Katja Filippova, Enrique Alfonseca, Carlos~A. Colmenares, Lukasz Kaiser, and
  Oriol Vinyals. 2015.
\newblock Sentence {{Compression}} by {{Deletion}} with {{LSTMs}}.
\newblock In \emph{Proceedings of the 2015 {{Conference}} on {{Empirical
  Methods}} in {{Natural Language Processing}}}, pages 360--368, {Lisbon,
  Portugal}. {Association for Computational Linguistics}.

\bibitem[{Filippova and Altun(2013)}]{filippova:2013}
Katja Filippova and Yasemin Altun. 2013.
\newblock {Overcoming the Lack of Parallel Data in Sentence Compression}.
\newblock In \emph{Proceedings of the 2013 Conference on Empirical Methods in
  Natural Language Processing}, pages 1481--1491, {Seattle, Washington, USA}.
  {Association for Computational Linguistics}.

\bibitem[{Fujita(2005)}]{fujita:2005}
Atsushi Fujita. 2005.
\newblock \emph{Automatic Generation of Syntactically Well-Formed and
  Semantically Appropriate Paraphrases}.
\newblock Ph.D. thesis, Nara Institute of Science and Technology.

\bibitem[{Ganitkevitch and {Callison-Burch}(2014)}]{ganitkevitch:2014}
Juri Ganitkevitch and Chris {Callison-Burch}. 2014.
\newblock {The Multilingual Paraphrase Database}.
\newblock In \emph{Proceedings of the Ninth International Conference on
  Language Resources and Evaluation ({{LREC}}'14)}, pages 4276--4283,
  {Reykjavik, Iceland}. {European Language Resources Association (ELRA)}.

\bibitem[{Ganitkevitch et~al.(2013)Ganitkevitch, Van~Durme, and
  {Callison-Burch}}]{ganitkevitch:2013}
Juri Ganitkevitch, Benjamin Van~Durme, and Chris {Callison-Burch}. 2013.
\newblock {{{PPDB}}: {{The}} Paraphrase Database}.
\newblock In \emph{Proceedings of the 2013 Conference of the North {{American}}
  Chapter of the Association for Computational Linguistics: {{Human}} Language
  Technologies}, pages 758--764, {Atlanta, Georgia}. {Association for
  Computational Linguistics}.

\bibitem[{Gao et~al.(2020)Gao, Zhang, Ou, and Yu}]{gao:2020}
Silin Gao, Yichi Zhang, Zhijian Ou, and Zhou Yu. 2020.
\newblock Paraphrase {{Augmented Task-Oriented Dialog Generation}}.
\newblock In \emph{Proceedings of the 58th {{Annual Meeting}} of the
  {{Association}} for {{Computational Linguistics}}}, pages 639--649, {Online}.
  {Association for Computational Linguistics}.

\bibitem[{Gilabert et~al.(2013)Gilabert, {Garcia-Mila}, and
  Felton}]{gilabert:2013}
Sandra Gilabert, Merce {Garcia-Mila}, and Mark~K. Felton. 2013.
\newblock The {{Effect}} of {{Task Instructions}} on {{Students}}' {{Use}} of
  {{Repetition}} in {{Argumentative Discourse}}.
\newblock \emph{International Journal of Science Education}, 35(17):2857--2878.

\bibitem[{Gohsen et~al.(2023)Gohsen, Hagen, Potthast, and Stein}]{gohsen:2023}
Marcel Gohsen, Matthias Hagen, Martin Potthast, and Benno Stein. 2023.
\newblock {Paraphrase Acquisition from Image Captions}.
\newblock In \emph{Proceedings of the 17th {{Conference}} of the {{European
  Chapter}} of the {{Association}} for {{Computational Linguistics}}}, pages
  3348--3358, {Dubrovnik, Croatia}. {Association for Computational
  Linguistics}.

\bibitem[{Goyal and Durrett(2020)}]{goyal:2020}
Tanya Goyal and Greg Durrett. 2020.
\newblock Neural {{Syntactic Preordering}} for {{Controlled Paraphrase
  Generation}}.
\newblock In \emph{Proceedings of the 58th {{Annual Meeting}} of the
  {{Association}} for {{Computational Linguistics}}}, pages 238--252, {Online}.
  {Association for Computational Linguistics}.

\bibitem[{Gupta et~al.(2018)Gupta, Agarwal, Singh, and Rai}]{gupta:2018}
Ankush Gupta, Arvind Agarwal, Prawaan Singh, and Piyush Rai. 2018.
\newblock A {{Deep Generative Framework}} for {{Paraphrase Generation}}.
\newblock \emph{Proceedings of the AAAI Conference on Artificial Intelligence},
  32(1).

\bibitem[{Gupta et~al.(2023)Gupta, V., Mohania, and Goyal}]{gupta:2023}
Rishabh Gupta, Venktesh V., Mukesh Mohania, and Vikram Goyal. 2023.
\newblock Coherence and {{Diversity}} through {{Noise}}: {{Self-Supervised
  Paraphrase Generation}} via {{Structure-Aware Denoising}}.

\bibitem[{Gurari et~al.(2020)Gurari, Zhao, Zhang, and
  Bhattacharya}]{gurari:2020}
Danna Gurari, Yinan Zhao, Meng Zhang, and Nilavra Bhattacharya. 2020.
\newblock Captioning {{Images Taken}} by {{People Who Are Blind}}.

\bibitem[{Hirst(2003)}]{hirst:2003}
Graeme Hirst. 2003.
\newblock {Paraphrasing Paraphrased.}

\bibitem[{Iyyer et~al.(2018)Iyyer, Wieting, Gimpel, and
  Zettlemoyer}]{iyyer:2018}
Mohit Iyyer, John Wieting, Kevin Gimpel, and Luke Zettlemoyer. 2018.
\newblock Adversarial {{Example Generation}} with {{Syntactically Controlled
  Paraphrase Networks}}.

\bibitem[{Jin et~al.(2022)Jin, Jin, Hu, Vechtomova, and Mihalcea}]{jin:2022}
Di~Jin, Zhijing Jin, Zhiting Hu, Olga Vechtomova, and Rada Mihalcea. 2022.
\newblock Deep {{Learning}} for {{Text Style Transfer}}: {{A Survey}}.
\newblock \emph{Computational Linguistics}, 48(1):155--205.

\bibitem[{Kauchak and Barzilay(2006)}]{kauchak:2006}
David Kauchak and Regina Barzilay. 2006.
\newblock Paraphrasing for automatic evaluation.
\newblock In \emph{Proceedings of the Human Language Technology Conference of
  the {{NAACL}}, Main Conference}, pages 455--462, {New York City, USA}.
  {Association for Computational Linguistics}.

\bibitem[{Khot et~al.(2018)Khot, Sabharwal, and Clark}]{khot:2018}
Tushar Khot, Ashish Sabharwal, and Peter Clark. 2018.
\newblock {{SciTaiL}}: {{A}} textual entailment dataset from science question
  answering.
\newblock In \emph{{{AAAI}} Conference on Artificial Intelligence}.

\bibitem[{Korman et~al.(2018)Korman, Mack, Jett, and Renear}]{korman:2018}
Daniel~Z. Korman, Eric Mack, Jacob Jett, and Allen~H. Renear. 2018.
\newblock {Defining Textual Entailment}.
\newblock \emph{Journal of the Association for Information Science and
  Technology}, 69(6):763--772.

\bibitem[{Krishna et~al.(2020)Krishna, Wieting, and Iyyer}]{krishna:2020}
Kalpesh Krishna, John Wieting, and Mohit Iyyer. 2020.
\newblock Reformulating {{Unsupervised Style Transfer}} as {{Paraphrase
  Generation}}.

\bibitem[{Kumar et~al.(2020)Kumar, Ahuja, Vadapalli, and Talukdar}]{kumar:2020}
Ashutosh Kumar, Kabir Ahuja, Raghuram Vadapalli, and Partha Talukdar. 2020.
\newblock Syntax-{{Guided Controlled Generation}} of {{Paraphrases}}.
\newblock \emph{Transactions of the Association for Computational Linguistics},
  8:330--345.

\bibitem[{Kumar et~al.(2019)Kumar, Bhattamishra, Bhandari, and
  Talukdar}]{kumar:2019}
Ashutosh Kumar, Satwik Bhattamishra, Manik Bhandari, and Partha Talukdar. 2019.
\newblock Submodular {{Optimization-based Diverse Paraphrasing}} and its
  {{Effectiveness}} in {{Data Augmentation}}.
\newblock In \emph{Proceedings of the 2019 {{Conference}} of the {{North}}},
  pages 3609--3619, {Minneapolis, Minnesota}. {Association for Computational
  Linguistics}.

\bibitem[{Lan et~al.(2017)Lan, Qiu, He, and Xu}]{lan:2017}
Wuwei Lan, Siyu Qiu, Hua He, and Wei Xu. 2017.
\newblock A {{Continuously Growing Dataset}} of {{Sentential Paraphrases}}.
\newblock In \emph{Proceedings of the 2017 {{Conference}} on {{Empirical
  Methods}} in {{Natural}} {{Language Processing}}}, pages 1224--1234,
  {Copenhagen, Denmark}. {Association for Computational Linguistics}.

\bibitem[{Li et~al.(2018)Li, Jiang, Shang, and Li}]{li:2018}
Zichao Li, Xin Jiang, Lifeng Shang, and Hang Li. 2018.
\newblock Paraphrase {{Generation}} with {{Deep Reinforcement Learning}}.
\newblock In \emph{Proceedings of the 2018 {{Conference}} on {{Empirical
  Methods}} in {{Natural Language Processing}}}, pages 3865--3878, {Brussels,
  Belgium}. {Association for Computational Linguistics}.

\bibitem[{Lin(2004)}]{lin:2004}
Chin-Yew Lin. 2004.
\newblock {{ROUGE}}: {{A}} package for automatic evaluation of summaries.
\newblock In \emph{Text Summarization Branches Out}, pages 74--81, {Barcelona,
  Spain}. {Association for Computational Linguistics}.

\bibitem[{Lin et~al.(2015)Lin, Maire, Belongie, Bourdev, Girshick, Hays,
  Perona, Ramanan, Zitnick, and Doll{\'a}r}]{lin:2015}
Tsung-Yi Lin, Michael Maire, Serge Belongie, Lubomir Bourdev, Ross Girshick,
  James Hays, Pietro Perona, Deva Ramanan, C.~Lawrence Zitnick, and Piotr
  Doll{\'a}r. 2015.
\newblock Microsoft {{COCO}}: {{Common Objects}} in {{Context}}.

\bibitem[{Logacheva et~al.(2022)Logacheva, Dementieva, Ustyantsev, Moskovskiy,
  Dale, Krotova, Semenov, and Panchenko}]{logacheva:2022}
Varvara Logacheva, Daryna Dementieva, Sergey Ustyantsev, Daniil Moskovskiy,
  David Dale, Irina Krotova, Nikita Semenov, and Alexander Panchenko. 2022.
\newblock {{ParaDetox}}: {{Detoxification}} with {{Parallel Data}}.
\newblock In \emph{Proceedings of the 60th {{Annual Meeting}} of the
  {{Association}} for {{Computational Linguistics}} ({{Volume}} 1: {{Long
  Papers}})}, pages 6804--6818, {Dublin, Ireland}. {Association for
  Computational Linguistics}.

\bibitem[{Lu and Lam(2023)}]{lu:2023}
Hongyuan Lu and Wai Lam. 2023.
\newblock {{PCC}}: {{Paraphrasing}} with bottom-k sampling and cyclic learning
  for curriculum data augmentation.
\newblock In \emph{Proceedings of the 17th Conference of the European Chapter
  of the Association for Computational Linguistics, {{EACL}} 2023, Dubrovnik,
  Croatia, May 2-6, 2023}, pages 68--82. {Association for Computational
  Linguistics}.

\bibitem[{Maddela et~al.(2021)Maddela, {Alva-Manchego}, and Xu}]{maddela:2021}
Mounica Maddela, Fernando {Alva-Manchego}, and Wei Xu. 2021.
\newblock Controllable {{Text Simplification}} with {{Explicit Paraphrasing}}.

\bibitem[{Madnani et~al.(2007)Madnani, Fazil~Ayan, Resnik, and
  Dorr}]{madnani:2007}
Nitin Madnani, Necip Fazil~Ayan, Philip Resnik, and Bonnie Dorr. 2007.
\newblock {Using Paraphrases for Parameter Tuning in Statistical Machine
  Translation}.
\newblock In \emph{Proceedings of the Second Workshop on Statistical Machine
  Translation}, pages 120--127, {Prague, Czech Republic}. {Association for
  Computational Linguistics}.

\bibitem[{Madnani et~al.(2008)Madnani, Resnik, Dorr, and
  Schwartz}]{madnani:2008}
Nitin Madnani, Philip Resnik, Bonnie~J. Dorr, and Richard Schwartz. 2008.
\newblock {Are Multiple Reference Translations Necessary? {{Investigating}} the
  Value of Paraphrased Reference Translations in Parameter Optimization}.
\newblock In \emph{Proceedings of the 8th Conference of the Association for
  Machine Translation in the Americas: {{Research}} Papers}, pages 143--152,
  {Waikiki, USA}. {Association for Machine Translation in the Americas}.

\bibitem[{McKeown(1983)}]{mckeown:1983}
Kathleen~R. McKeown. 1983.
\newblock {Paraphrasing Questions Using given and New Information}.
\newblock \emph{American Journal of Computational Linguistics}, 9(1):1--10.

\bibitem[{Murata and Isahara(2001)}]{murata:2001}
Masaki Murata and Hitoshi Isahara. 2001.
\newblock Universal {{Model}} for {{Paraphrasing}} -- {{Using Transformation
  Based}} on a {{Defined Criteria}} --.

\bibitem[{Owczarzak et~al.(2006)Owczarzak, Groves, Van~Genabith, and
  Way}]{owczarzak:2006}
Karolina Owczarzak, Declan Groves, Josef Van~Genabith, and Andy Way. 2006.
\newblock {Contextual Bitext-Derived Paraphrases in Automatic {{MT}}
  Evaluation}.
\newblock In \emph{Proceedings on the Workshop on Statistical Machine
  Translation}, pages 86--93, {New York City}. {Association for Computational
  Linguistics}.

\bibitem[{Papineni et~al.(2002)Papineni, Roukos, Ward, and Zhu}]{papineni:2002}
Kishore Papineni, Salim Roukos, Todd Ward, and Wei-Jing Zhu. 2002.
\newblock {Bleu: A Method for Automatic Evaluation of Machine Translation}.
\newblock In \emph{Proceedings of the 40th Annual Meeting of the Association
  for Computational Linguistics}, pages 311--318, {Philadelphia, Pennsylvania,
  USA}. {Association for Computational Linguistics}.

\bibitem[{Pavlick et~al.(2015)Pavlick, Rastogi, Ganitkevitch, Van~Durme, and
  {Callison-Burch}}]{pavlick:2015}
Ellie Pavlick, Pushpendre Rastogi, Juri Ganitkevitch, Benjamin Van~Durme, and
  Chris {Callison-Burch}. 2015.
\newblock {{PPDB}} 2.0: {{Better}} paraphrase ranking, fine-grained entailment
  relations, word embeddings, and style classification.
\newblock In \emph{Proceedings of the 53rd {{Annual Meeting}} of the
  {{Association}} for {{Computational Linguistics}} and the 7th {{International
  Joint Conference}} on {{Natural Language Processing}} ({{Volume}} 2: {{Short
  Papers}})}, pages 425--430, {Beijing, China}. {Association for Computational
  Linguistics}.

\bibitem[{Potthast et~al.(2010)Potthast, {Barr{\'o}n-Cede{\~n}o}, Eiselt,
  Stein, and Rosso}]{potthast:2010}
Martin Potthast, Alberto {Barr{\'o}n-Cede{\~n}o}, Andreas Eiselt, Benno Stein,
  and Paolo Rosso. 2010.
\newblock {Overview of the 2nd International Competition on Plagiarism
  Detection}.
\newblock In \emph{Working Notes Papers of the {{CLEF}} 2010 Evaluation Labs},
  volume 1176 of \emph{Lecture Notes in Computer Science}.

\bibitem[{Prakash et~al.(2016)Prakash, Hasan, Lee, Datla, Qadir, Liu, and
  Farri}]{prakash:2016}
Aaditya Prakash, Sadid~A. Hasan, Kathy Lee, Vivek~V. Datla, Ashequl Qadir, Joey
  Liu, and Oladimeji Farri. 2016.
\newblock {Neural Paraphrase Generation with Stacked Residual {{LSTM}}
  Networks}.

\bibitem[{Qian et~al.(2019)Qian, Qiu, Zhang, Jiang, and Yu}]{qian:2019}
Lihua Qian, Lin Qiu, Weinan Zhang, Xin Jiang, and Yong Yu. 2019.
\newblock {Exploring Diverse Expressions for Paraphrase Generation}.
\newblock In \emph{Proceedings of the 2019 Conference on Empirical Methods in
  Natural Language Processing and the 9th International Joint Conference on
  Natural Language Processing ({{EMNLP-IJCNLP}})}, pages 3173--3182, {Hong
  Kong, China}. {Association for Computational Linguistics}.

\bibitem[{Qiu et~al.(2023)Qiu, Chen, and Yu}]{qiu:2023}
Dong Qiu, Lei Chen, and Yang Yu. 2023.
\newblock {Document-Level Paraphrase Generation Base on Attention Enhanced
  Graph {{LSTM}}}.
\newblock \emph{Applied Intelligence}, 53(9):10459--10471.

\bibitem[{{Quora}(2017)}]{quora:2017}
{Quora}. 2017.
\newblock Quora {{Question Pairs}}.

\bibitem[{Reimers and Gurevych(2019)}]{reimers:2019}
Nils Reimers and Iryna Gurevych. 2019.
\newblock Sentence-{{BERT}}: {{Sentence Embeddings}} using {{Siamese
  BERT-Networks}}.

\bibitem[{Rush et~al.(2015)Rush, Chopra, and Weston}]{rush:2015}
Alexander~M. Rush, Sumit Chopra, and Jason Weston. 2015.
\newblock A {{Neural Attention Model}} for {{Abstractive Sentence
  Summarization}}.
\newblock In \emph{Proceedings of the 2015 {{Conference}} on {{Empirical
  Methods}} in {{Natural Language Processing}}}, pages 379--389, {Lisbon,
  Portugal}. {Association for Computational Linguistics}.

\bibitem[{Safovich and Azaria(2020)}]{safovich:2020}
Yuri Safovich and Amos Azaria. 2020.
\newblock Fiction {{Sentence Expansion}} and {{Enhancement}} via {{Focused
  Objective}} and {{Novelty Curve Sampling}}.
\newblock In \emph{2020 {{IEEE}} 32nd {{International Conference}} on {{Tools}}
  with {{Artificial Intelligence}} ({{ICTAI}})}, pages 835--843, {Baltimore,
  MD, USA}. {IEEE}.

\bibitem[{Scherrer(2020)}]{scherrer:2020}
Yves Scherrer. 2020.
\newblock {{{TaPaCo}}: {{A}} Corpus of Sentential Paraphrases for 73
  Languages}.
\newblock In \emph{Proceedings of the Twelfth Language Resources and Evaluation
  Conference}, pages 6868--6873, {Marseille, France}. {European Language
  Resources Association}.

\bibitem[{Schlangen(2004)}]{schlangen:2004}
David Schlangen. 2004.
\newblock {Causes and Strategies for Requesting Clarification in Dialogue}.
\newblock In \emph{Proceedings of the 5th {{SIGdial Workshop}} on {{Discourse}}
  and {{Dialogue}} at {{HLT-NAACL}} 2004}, pages 136--143, {Cambridge,
  Massachusetts, USA}. {Association for Computational Linguistics}.

\bibitem[{Sordoni et~al.(2015)Sordoni, Bengio, Vahabi, Lioma, Grue~Simonsen,
  and Nie}]{sordoni:2015}
Alessandro Sordoni, Yoshua Bengio, Hossein Vahabi, Christina Lioma, Jakob
  Grue~Simonsen, and Jian-Yun Nie. 2015.
\newblock A {{Hierarchical Recurrent Encoder-Decoder}} for {{Generative
  Context-Aware Query Suggestion}}.
\newblock In \emph{Proceedings of the 24th {{ACM International}} on
  {{Conference}} on {{Information}} and {{Knowledge Management}}}, pages
  553--562, {Melbourne Australia}. {ACM}.

\bibitem[{Stein et~al.(2014)Stein, Hagen, and Br{\"a}utigam}]{stein:2014}
Benno Stein, Matthias Hagen, and Christof Br{\"a}utigam. 2014.
\newblock {Generating Acrostics via Paraphrasing and Heuristic Search}.
\newblock In \emph{Proceedings of {{COLING}} 2014, the 25th International
  Conference on Computational Linguistics: {{Technical}} Papers}, pages
  2018--2029, {Dublin, Ireland}. {Dublin City University and Association for
  Computational Linguistics}.

\bibitem[{Sun and Zhou(2012)}]{sun:2012}
Hong Sun and Ming Zhou. 2012.
\newblock {Joint Learning of a Dual {{SMT}} System for Paraphrase Generation}.
\newblock In \emph{Proceedings of the 50th Annual Meeting of the Association
  for Computational Linguistics (Volume 2: {{Short}} Papers)}, pages 38--42,
  {Jeju Island, Korea}. {Association for Computational Linguistics}.

\bibitem[{Szegedy et~al.(2014)Szegedy, Zaremba, Sutskever, Bruna, Erhan,
  Goodfellow, and Fergus}]{szegedy:2014}
Christian Szegedy, Wojciech Zaremba, Ilya Sutskever, Joan Bruna, Dumitru Erhan,
  Ian Goodfellow, and Rob Fergus. 2014.
\newblock {Intriguing Properties of Neural Networks}.

\bibitem[{Topkara et~al.(2005)Topkara, Taskiran, and Delp~Iii}]{topkara:2005}
Mercan Topkara, Cuneyt~M. Taskiran, and Edward~J. Delp~Iii. 2005.
\newblock Natural language watermarking.
\newblock In \emph{Electronic {{Imaging}} 2005}, page 441, {San Jose, CA}.

\bibitem[{Toutanova et~al.(2016)Toutanova, Brockett, Tran, and
  Amershi}]{toutanova:2016}
Kristina Toutanova, Chris Brockett, Ke~M. Tran, and Saleema Amershi. 2016.
\newblock A {{Dataset}} and {{Evaluation Metrics}} for {{Abstractive
  Compression}} of {{Sentences}} and {{Short Paragraphs}}.
\newblock In \emph{Proceedings of the 2016 {{Conference}} on {{Empirical
  Methods}} in {{Natural}} {{Language Processing}}}, pages 340--350, {Austin,
  Texas}. {Association for Computational Linguistics}.

\bibitem[{Vila et~al.(2014)Vila, Mart{\'i}, and Rodr{\'i}guez}]{vila:2014}
Marta Vila, M.~Ant{\`o}nia Mart{\'i}, and Horacio Rodr{\'i}guez. 2014.
\newblock Is {{This}} a {{Paraphrase}}? {{What Kind}}? {{Paraphrase
  Boundaries}} and {{Typology}}.
\newblock \emph{Open Journal of Modern Linguistics}, 04(01):205--218.

\bibitem[{Wahle et~al.(2022{\natexlab{a}})Wahle, Ruas, Folt{\'y}nek, Meuschke,
  and Gipp}]{wahle:2022a}
Jan~Philip Wahle, Terry Ruas, Tom{\'a}{\v s} Folt{\'y}nek, Norman Meuschke, and
  Bela Gipp. 2022{\natexlab{a}}.
\newblock {Identifying Machine-Paraphrased Plagiarism}.
\newblock In \emph{Information for a Better World: {{Shaping}} the Global
  Future}, pages 393--413, {Cham}. {Springer International Publishing}.

\bibitem[{Wahle et~al.(2022{\natexlab{b}})Wahle, Ruas, Kirstein, and
  Gipp}]{wahle:2022}
Jan~Philip Wahle, Terry Ruas, Frederic Kirstein, and Bela Gipp.
  2022{\natexlab{b}}.
\newblock {How Large Language Models Are Transforming Machine-Paraphrase
  Plagiarism}.
\newblock In \emph{Proceedings of the 2022 Conference on Empirical Methods in
  Natural Language Processing}, pages 952--963, {Abu Dhabi, United Arab
  Emirates}. {Association for Computational Linguistics}.

\bibitem[{Wang et~al.(2019)Wang, Gupta, Chang, and Baldridge}]{wang:2019}
Su~Wang, Rahul Gupta, Nancy Chang, and Jason Baldridge. 2019.
\newblock A {{Task}} in a {{Suit}} and a {{Tie}}: {{Paraphrase Generation}}
  with {{Semantic Augmentation}}.
\newblock \emph{Proceedings of the AAAI Conference on Artificial Intelligence},
  33(01):7176--7183.

\bibitem[{Wang and Yang(2015)}]{wang:2015}
William~Yang Wang and Diyi Yang. 2015.
\newblock That's {{So Annoying}}!!!: {{A Lexical}} and {{Frame-Semantic
  Embedding Based Data Augmentation Approach}} to {{Automatic Categorization}}
  of {{Annoying Behaviors}} using \#petpeeve {{Tweets}}.
\newblock In \emph{Proceedings of the 2015 {{Conference}} on {{Empirical
  Methods}} in {{Natural Language Processing}}}, pages 2557--2563, {Lisbon,
  Portugal}. {Association for Computational Linguistics}.

\bibitem[{Wieting and Gimpel(2018)}]{wieting:2018}
John Wieting and Kevin Gimpel. 2018.
\newblock {{ParaNMT-50M}}: {{Pushing}} the {{Limits}} of {{Paraphrastic
  Sentence Embeddings}} with {{Millions}} of {{Machine Translations}}.
\newblock In \emph{Proceedings of the 56th {{Annual Meeting}} of the
  {{Association}} for {{Computational Linguistics}} ({{Volume}} 1: {{Long
  Papers}})}, pages 451--462, {Melbourne, Australia}. {Association for
  Computational Linguistics}.

\bibitem[{Wilson et~al.(2014)Wilson, Blunsom, and Ker}]{wilson:2014}
Alex Wilson, Phil Blunsom, and Andrew~D. Ker. 2014.
\newblock Linguistic steganography on {{Twitter}}: Hierarchical language
  modeling with manual interaction.
\newblock In \emph{{{IS}}\&{{T}}/{{SPIE Electronic Imaging}}}, page 902803,
  {San Francisco, California, USA}.

\bibitem[{Wubben et~al.(2010)Wubben, {van den Bosch}, and
  Krahmer}]{wubben:2010}
Sander Wubben, Antal {van den Bosch}, and Emiel Krahmer. 2010.
\newblock {Paraphrase Generation as Monolingual Translation: {{Data}} and
  Evaluation}.
\newblock In \emph{Proceedings of the 6th International Natural Language
  Generation Conference}. {Association for Computational Linguistics}.

\bibitem[{Xu et~al.(2021)Xu, Lu, Sun, Ma, and Guo}]{xu:2021}
Haoran Xu, Sixing Lu, Zhongkai Sun, Chengyuan Ma, and Chenlei Guo. 2021.
\newblock {{VAE}} based {{Text Style Transfer}} with {{Pivot Words Enhancement
  Learning}}.

\bibitem[{Xu et~al.(2016)Xu, Napoles, Pavlick, Chen, and
  {Callison-Burch}}]{xu:2016}
Wei Xu, Courtney Napoles, Ellie Pavlick, Quanze Chen, and Chris
  {Callison-Burch}. 2016.
\newblock Optimizing {{Statistical Machine Translation}} for {{Text
  Simplification}}.
\newblock \emph{Transactions of the Association for Computational Linguistics},
  4:401--415.

\bibitem[{Yanaka et~al.(2019)Yanaka, Mineshima, Bekki, Inui, Sekine,
  Abzianidze, and Bos}]{yanaka:2019}
Hitomi Yanaka, Koji Mineshima, Daisuke Bekki, Kentaro Inui, Satoshi Sekine,
  Lasha Abzianidze, and Johan Bos. 2019.
\newblock {{HELP}}: {{A Dataset}} for {{Identifying Shortcomings}} of {{Neural
  Models}} in {{Monotonicity Reasoning}}.

\bibitem[{Yang et~al.(2021)Yang, Halfaker, Kraut, and Hovy}]{yang:2021}
Diyi Yang, Aaron Halfaker, Robert Kraut, and Eduard Hovy. 2021.
\newblock Who {{Did What}}: {{Editor Role Identification}} in {{Wikipedia}}.
\newblock \emph{Proceedings of the International AAAI Conference on Web and
  Social Media}, 10(1):446--455.

\bibitem[{Yimam and Biemann(2018)}]{yimam:2018}
Seid~Muhie Yimam and Chris Biemann. 2018.
\newblock {{Par4Sim}} -- {{Adaptive Paraphrasing}} for {{Text Simplification}}.

\bibitem[{Young et~al.(1970)Young, Becker, and Pike}]{young:1970}
Richard~Emerson Young, Alton~L. Becker, and Kenneth~L. Pike. 1970.
\newblock \emph{Rhetoric: Discovery and Change}.
\newblock {Harcourt, Brace \& World}, {New York}.

\bibitem[{Yu et~al.(2018)Yu, Dohan, Luong, Zhao, Chen, Norouzi, and
  Le}]{yu:2018}
Adams~Wei Yu, David Dohan, Minh-Thang Luong, Rui Zhao, Kai Chen, Mohammad
  Norouzi, and Quoc~V. Le. 2018.
\newblock {{QANet}}: {{Combining Local Convolution}} with {{Global
  Self-Attention}} for {{Reading Comprehension}}.

\bibitem[{Zhang et~al.(2016)Zhang, Zhao, and LeCun}]{zhang:2016}
Xiang Zhang, Junbo Zhao, and Yann LeCun. 2016.
\newblock Character-level {{Convolutional Networks}} for {{Text
  Classification}}.

\bibitem[{Zhang and Lapata(2017)}]{zhang:2017}
Xingxing Zhang and Mirella Lapata. 2017.
\newblock Sentence {{Simplification}} with {{Deep Reinforcement Learning}}.
\newblock In \emph{Proceedings of the 2017 {{Conference}} on {{Empirical
  Methods}} in {{Natural}} {{Language Processing}}}, pages 584--594,
  {Copenhagen, Denmark}. {Association for Computational Linguistics}.

\bibitem[{Zhang et~al.(2019)Zhang, Baldridge, and He}]{zhang:2019}
Yuan Zhang, Jason Baldridge, and Luheng He. 2019.
\newblock {{PAWS}}: {{Paraphrase Adversaries}} from {{Word Scrambling}}.

\bibitem[{Zhao et~al.(2009)Zhao, Lan, Liu, and Li}]{zhao:2009}
Shiqi Zhao, Xiang Lan, Ting Liu, and Sheng Li. 2009.
\newblock Application-driven statistical paraphrase generation.
\newblock In \emph{Proceedings of the Joint Conference of the 47th Annual
  Meeting of the {{ACL}} and the 4th International Joint Conference on Natural
  Language Processing of the {{AFNLP}}}, pages 834--842, {Suntec, Singapore}.
  {Association for Computational Linguistics}.

\bibitem[{Zhou and Bhat(2021)}]{zhou:2021}
Jianing Zhou and Suma Bhat. 2021.
\newblock Paraphrase {{Generation}}: {{A Survey}} of the {{State}} of the
  {{Art}}.
\newblock In \emph{Proceedings of the 2021 {{Conference}} on {{Empirical
  Methods}} in {{Natural Language Processing}}}, pages 5075--5086, {Online and
  Punta Cana, Dominican Republic}. {Association for Computational Linguistics}.

\bibitem[{Ziegler et~al.(2019)Ziegler, Deng, and Rush}]{ziegler:2019}
Zachary~M. Ziegler, Yuntian Deng, and Alexander~M. Rush. 2019.
\newblock Neural {{Linguistic Steganography}}.

\bibitem[{Ziems et~al.(2022)Ziems, Li, Zhang, and Yang}]{ziems:2022}
Caleb Ziems, Minzhi Li, Anthony Zhang, and Diyi Yang. 2022.
\newblock Inducing {{Positive Perspectives}} with {{Text Reframing}}.

\end{thebibliography}



\end{document}